\documentclass[runningheads,a4paper]{llncs}

\usepackage{amssymb}
\usepackage{graphicx}
\usepackage{lineno,hyperref}
\usepackage{enumitem}
\usepackage[marginal]{footmisc}
\usepackage{subcaption}
\usepackage{amsmath}
\usepackage{multirow}
\usepackage{array}
\usepackage{tikz}
\usepackage{wrapfig}

\usepackage{url}

\begin{document}

\mainmatter  

\title{Learning Crisp Edge Detector \\ Using Logical Refinement Network}


%
%
\author{Luyan Liu$^{1}$, Kai Ma$^1$, Yefeng Zheng$^{1}$}
\institute{$^1$Tencent Jarvis Lab, Shenzhen, China}
%


%
%

\toctitle{Lecture Notes in Computer Science}
\tocauthor{Authors' Instructions}
\maketitle

\begin{abstract}
Edge detection is a fundamental problem in different computer vision tasks. Recently, edge detection algorithms achieve satisfying improvement built upon deep learning. Although most of them report favorable evaluation scores, they often fail to accurately localize edges and give thick and blurry boundaries. In addition, most of them focus on 2D images and the challenging 3D edge detection is still under-explored. In this work, we propose a novel logical refinement network for crisp edge detection, which is motivated by the logical relationship between segmentation and edge maps and can be applied to both 2D and 3D images. The network consists of a joint object and edge detection network and a crisp edge refinement network, which predicts more accurate, clearer and thinner high quality binary edge maps without any post-processing. Extensive experiments are conducted on the 2D nuclei images from Kaggle 2018 Data Science Bowl and a private 3D microscopy images of a monkey brain, which show outstanding performance compared with state-of-the-art methods.
\end{abstract}

\section{Introduction}
Edge detection is a fundamental and important task in computer vision which provides ample supplementary cues to other tasks like semantic and instance segmentation \cite{zimmermann2019faster}, object detection \cite{Alpher04} and recognition \cite{girshick2014rich}, scene understanding \cite{Alpher20}, etc. Until now, there are numerous works about edge detection. Most of the early works usually detect edges based on color, intensity, texture \cite{Alpher10} and other low-level learning features \cite{Alpher11}. Benefiting from the recent prevailing trend of deep learning algorithms, the edge detection methods have been greatly improved from the earlier traditional handcrafted feature based methods \cite{Alpher10,dollar2013structured}, to the deep learning based methods \cite{liu2017richer1,liu2019richer,xie15hed}. DeepEdge \cite{Alpher14}, DeepContour \cite{Alpher15} and DeepCrisp \cite{Alpher16} were proposed to leverage deep convolutional neural networks to facilitate the exploration of more discriminative features for edge detection. Bertasius et al. \cite{Alpher17} showed that the low-level boundary detection process could benefit from the high-level object features. A simultaneous edge alignment and learning method \cite{yu2018seal} was proposed to improve edge quality by addressing the label misalignment problem. Hierarchical multiscale and multilevel information was exploited in \cite{Qin_2019_CVPR,he2019bdcn,liu2017richer1,liu2019richer,xie15hed} to make full use of rich features for edge detection. Even though the existing methods are propitious to generate edge maps with high scores, they may confront problems of having low quality edge maps due to blurry and deviating from actual image boundaries \cite{Alpher16,hu2019dynamic}. The shortcoming may be adversarial for tasks required crisper and sharper boundaries like optical flow and image segmentation. Additionally, and importantly, most of the recent methods are designed for 2D images, leaving the 3D edge detection an open problem. In this paper, we propose a novel coarse-refine framework to address the challenging problems above, which can be applied to both 2D and 3D images.

\begin{figure}[t] 
	\centering\includegraphics[width=1\linewidth]{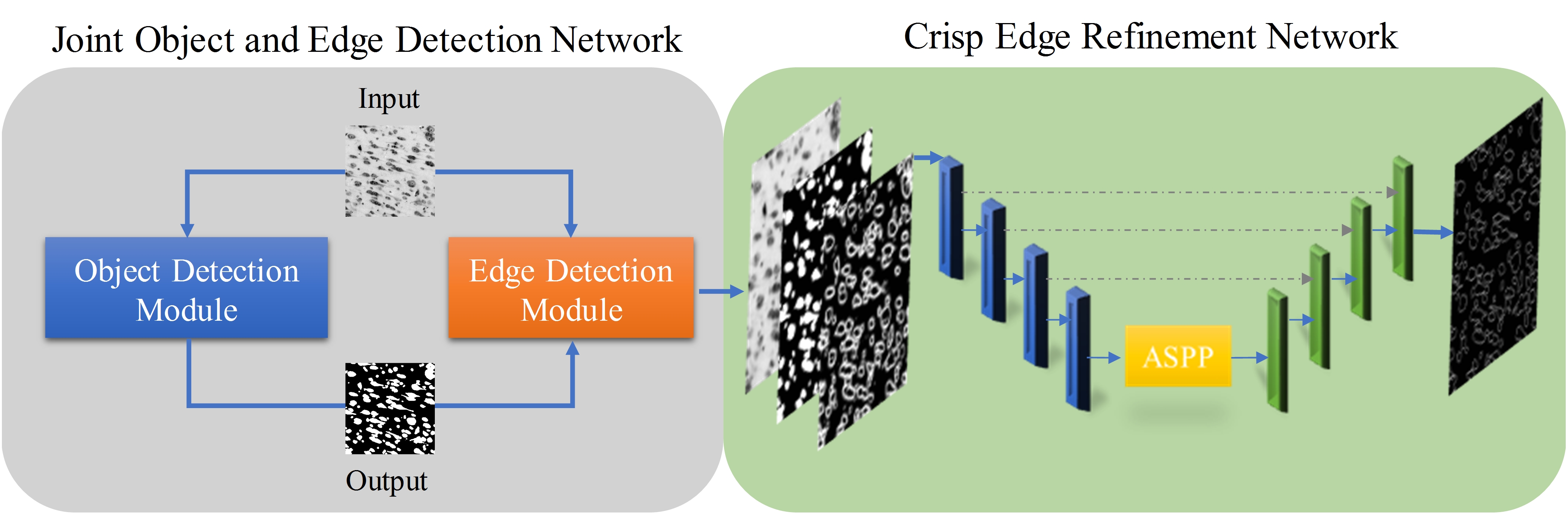} 
	\caption{The framework of our proposed logical refinement network.}\label{fig: 3DEdge-Net} 
\end{figure}

By exploring the relationship between binary segmentation and edge maps, it is not difficult to find out that the segmentation map contains ample edge information and the edge map is a subset of segmentation. Inspired by this observation, we propose a novel logical refinement network and a novel logical gate to generate accurate and crisp edges. The logical gate leverages the relationship between edge and segmentation maps to refine edge features. When it works in conjunction with logical refinement network, the quality of edge maps can be gradually improved. Our contributions can be summarized into the following two parts: 1) We propose a simple yet efficient logical gate, which passes feature maps between object and edge detection tasks, to simultaneously refine edge features. 2) We propose a novel coarse-refine framework for crisp edge detection, named logical refinement network, which utilizes the interrelation between object and edge maps to detect crisper and thinner image edges. Extensive experiments show that our proposed framework is effective on both 2D and 3D images.

\begin{figure}[t]
	\begin{center}
		\includegraphics[width=1\linewidth,height=.3\linewidth]{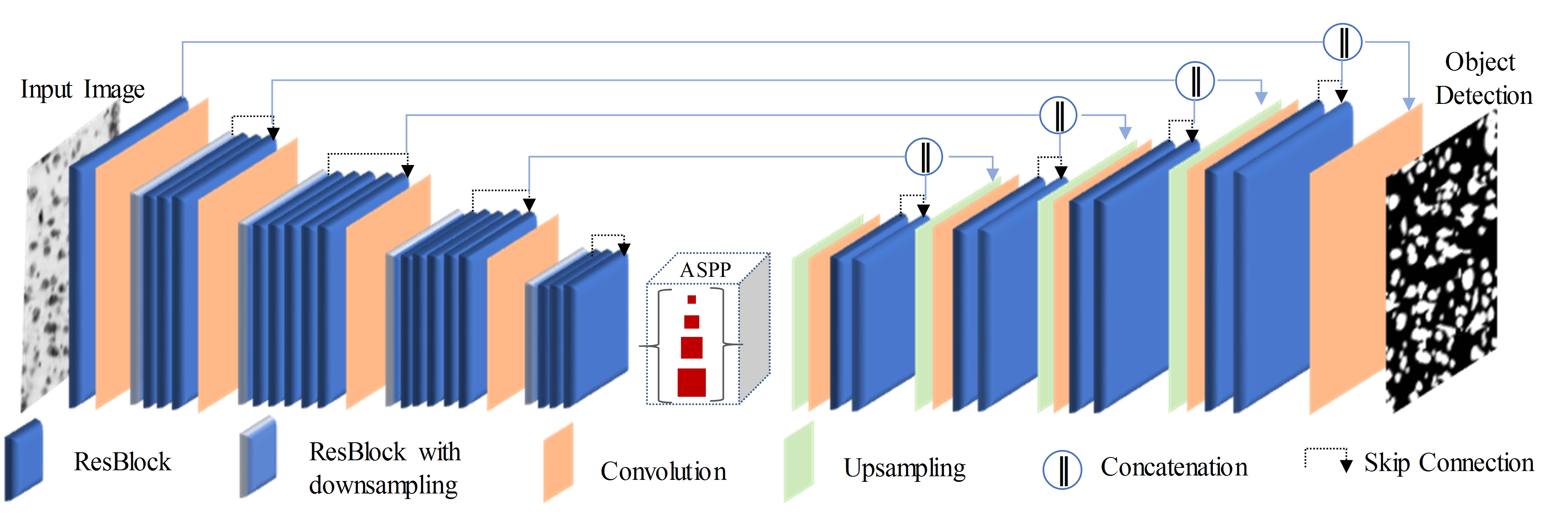}
	\end{center}
	\caption{The architecture of the object detection module (cf. the blue box in Fig. 1).}
	\label{fig:long}
	\label{fig:onecol}
\end{figure}


\section{Methodology}\label{Method}
Our method leverages a coarse-refine manner to capture both coarse (global) and fine (local) information to produce finer image boundaries, and it includes two phases: the first phase is to detect coarse edge maps using a joint object and edge detection network, while the second phase is to refine edge maps with the edge refinement network. Fig. 1 shows the framework of our proposed method. In the first phase, object and edge detection modules, which are shown in Figs. 2 and 3 respectively, are connected with logical gate and jointly trained end-to-end. The object information and coarse edge maps obtained in the first phase are then fed into the second phase to refine the final edge maps. The logical gate is also used in the second phase to gradually refine the edge features. In this section, we first introduce the logical gate, and then illustrate the work mechanism of the framework.

\noindent\textbf{Logical Gate.} \label{sec:logical gate}
Both edge and object detection can be treated as a pixel-wise binary classification task. An object detection map can be defined as $D_O$ that highlights the full semantic areas of the target object. Then, the edge detection map can be defined as $D_E$, which highlights the edge of objects only. Under this definition, the pixels in $D_E$ belong to a subset of the pixels in $D_O$, in which the logical relationship between edge and object maps can be reformulated as $D_E$ $\cap$ $D_O$=$D_E$, and $D_E$ $\cup$ $D_O$=$D_O$, where $\cap$ is Boolean AND operation and $\cup$ is Boolean OR operation. In this paper, we exploit such interrelation with a logical gate to refine edge features.

The object ($D_O$) and edge ($D_E$) maps are integrated by the logical gate operation $G$ (cf. the light-green box in Figs. 3 and 4), which is formulated as $G$=$Conv(D_E \bigoplus (D_O\bigotimes D_E))$, where $\bigoplus$ is element-wise addition, $\bigotimes$ is element-wise multiplication and $Conv$ is a $3 \times 3$ convolutional layer. In Fig. 3, the feature map $F$ also plays a role as edge map, since there are plenty of edge contexts contained in $F$. In Fig. 4, the object and edge maps obtained in the first phase are used as $D_O$ and $D_E$ to refine final edge maps. After applying the logical gate operation in the networks, the edge features will become clearer and crisper. It is because that the object features contain complete edge information and can be utilized to improve edge features by multiplication operation, and the distractors in the segmentation features can be suppressed by adding the edge features.

\begin{figure}[t]
	\begin{center}
		\includegraphics[width=1\linewidth,height=.35\linewidth]{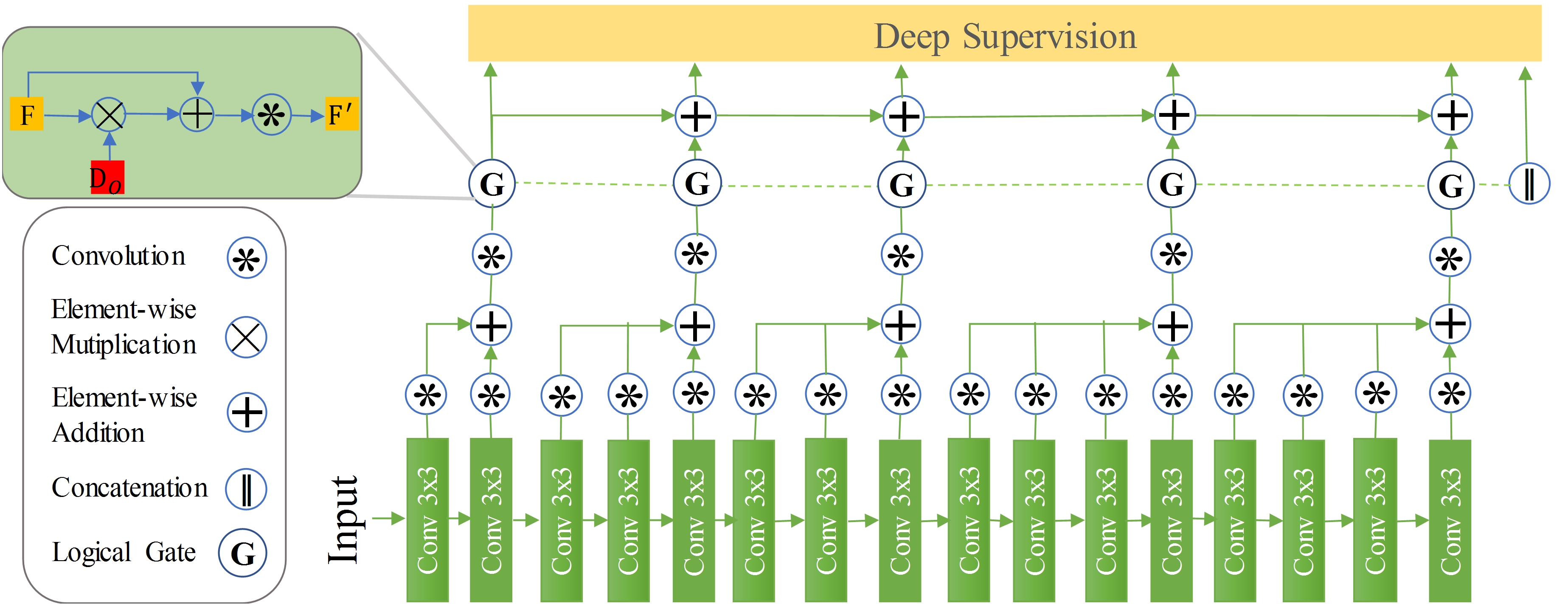}
	\end{center}
	\caption{The architecture of the edge detection module (cf. the orange box in Fig. 1).}
	\label{fig:long}
	\label{fig:onecol}
\end{figure}

\noindent\textbf{Joint Object and Edge Detection  Network.} \label{sec:2D-Net}
The joint object and edge detection network includes edge and object detection modules. Inspired by U-Net \cite{ronneberger2015u} and SegNet \cite{badrinarayanan2017segnet}, the object detection module is designed as an encoder-decoder architecture, which is shown in Fig. 2, since this kind of architecture can capture high level global contexts and low level local details at the same time. The encoder part has an input residual convolution block and four stages with 4, 6, 6, 4 residual blocks, respectively. The input and output of each stage is element-wise added in skip connection. To further capture local details and explore sufficient multiscale features, an Atrous Spatial Pyramid Pooling (ASPP)  block is used after the last stage of the encoder. The decoder includes four stages and an output convolutional filter, and there are two residual blocks in each stage. The output of object detection module can be defined as $D_O$, which is exploited by the logical gate in the edge detection module to improve the edge features in the first phase.

The 2D edge detection module is a simple yet efficient network, which is shown in Fig. 3. The network is composed of 16 convolutional layers which are divided into five stages. Each convolutional layer is connected to another convolutional layer and all resulting feature maps are element-wise accumulated to fuse feature maps into five stages. Each stage is followed by a convolutional layer. The feature map ($F$) after the convolutional layer, which also plays a role as edge map ($D_E$), is fed into logical gate $G$ along with the object map ($D_O$) generated from the object detection module. Specifically, the input feature map will be element-wise multiplied by the object map ($D_O$) and then element-wise added by the edge map ($D_E$), after which a convolutional layer is followed. The output of each logical gate operation ($F'$) is layer-specifically and deeply supervised. The obtained edge and object maps are then passed to the next phase to refine the coarse and thick edge maps obtained in this phase.

\begin{figure}[t]
	\begin{center}
		\includegraphics[width=1\linewidth,height=.35\linewidth]{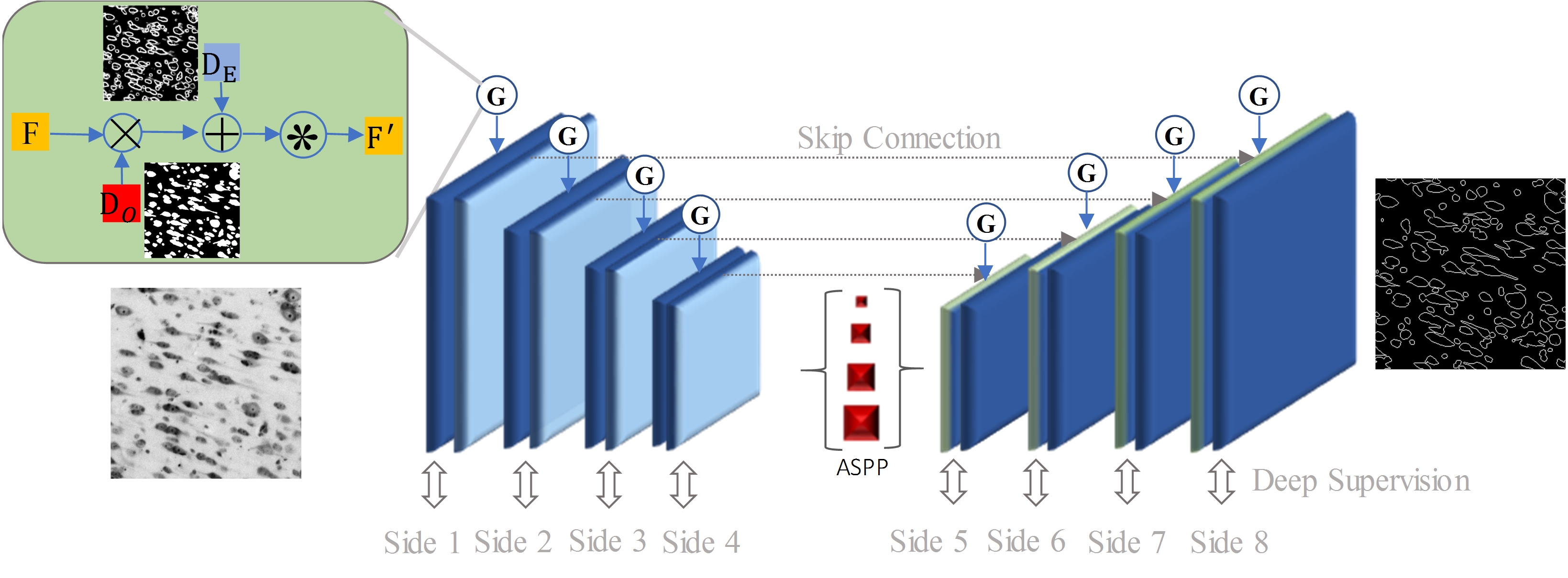}
	\end{center}
	\caption{The architecture of the crisp edge refinement network (cf. the green box in Fig. 1).}
	\label{fig:long}
	\label{fig:onecol}
\end{figure}

\noindent\textbf{Crisp Edge Refinement Network.} \label{sec:HUNet}
 The edge maps obtained from the first phase is usually coarse and unsharp. Such ``coarse'' appears in the following two aspects: one is the blurry and noisy boundaries, and the other one is the unevenly predicted probabilities. To address drawbacks in the edge maps, we develop a novel crisp edge refinement network with an encoder-decoder architecture, which is shown in Fig. 4. Both encoder and decoder have four stages. Each stage has two convolutional layers followed by ReLU  and group normalization. In the last stage of the encoder, an ASPP block is also added to enrich hierarchical features. The output feature map ($F$) of each stage is passed to logical gate G with edge ($D_E$) and object ($D_O$) maps obtained from the first phase, and the updated feature map ($F'$) is input to the next stage. All stages are fused together and thus result in nine outputs, which are deeply supervised by ground truth.

\noindent\textbf{Hybrid Loss Function.} \label{sec:HUNet}
The object detection module is trained with a cross-entropy loss. To address the class imbalance problem in edge detection and obtain high quality edge maps, we define a hybrid loss, including a focal loss and a cross-entropy loss, as the training loss of the edge detection module and the crisp edge refinement network. It is defined as the summation over all outputs:
\begin{align}
	\mathcal{L}=\sum_{k=1}^K{\alpha_k\ell^k}=\sum_{k=1}^K{\alpha_k(\ell_{ce}^k + \ell_{focal}^k)}
\end{align}
where $\ell^k$ is the loss of $k$-th side output, $K$ is the total number of outputs, and $\alpha_k$ is the weight of each loss. As described above, the edge detection module and crisp edge refinement network are deeply supervised with six (K=6) and nine (K=9) outputs, respectively.

\begin{table*}[t]
	\setlength{\abovecaptionskip}{0pt}
	\setlength{\belowcaptionskip}{10pt}
	\renewcommand{\arraystretch}{1.0}
	\centering
	\caption{Evaluation of the effectiveness of the proposed logical gate.}\label{table:JointLearningGate}
	\begin{tabular}{|c|c|c|c|c||c|c|c|c|}
		\hline
		\multirow{2}{*} & \multicolumn{4}{c||}{Phase 1} & \multicolumn{4}{c|}{Phase 2}   \\ \cline{2-9} 
		& ODS (\%)  &   OIS (\%)    & DSC (\%)  & HD  & ODS (\%)  & OIS (\%) & DSC (\%) & HD  \\ \hline
		{W/O G} & 80.84   &  81.20     & 35.23   & 22.298    &  - &   -  &   -   & -
		\\ \hline
		{With G} &  82.11    &  83.52    & 32.08   & 18.572  &  85.05 & 85.05  & 50.27  & 12.811
		\\ \hline
	\end{tabular}
\end{table*}

\begin{figure}[t]
	\begin{center}
		\includegraphics[width=1\linewidth]{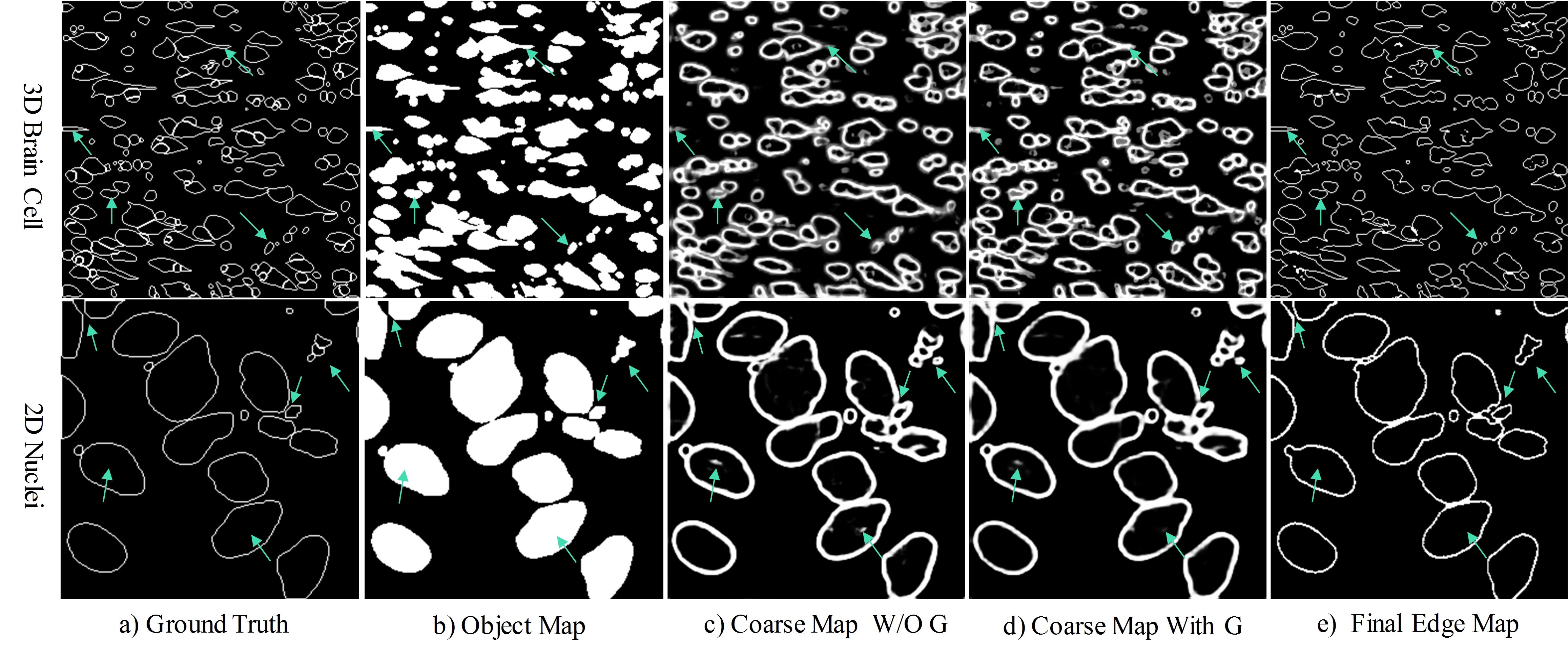}
	\end{center}
	\caption{The evaluation results with and without the proposed logical gate (Coarse Map: edge map predicted in the first phase).}
	\label{fig:long}
	\label{fig:onecol}
\end{figure}

\section{Experiments}
\noindent\textbf{Datasets and Implementation Details.} 
We evaluate our method on a 3D monkey brain cell dataset acquired from a special 3D light-sheet microscopy imaging equipment and a 2D nuclei detection dataset from Kaggle 2018 Data Science Bowl \cite{caicedo2019nucleus}. The 3D dataset contains six patches with size of $300\times300\times100$ voxels sampled from different regions of monkey brain images. The 2D dataset includes 634 images from training set of Kaggle 2018 Data Science Bowl which are resized to $256\times256$ pixels. Both datasets are originally designed for instance segmentation and very challenging since most of the images contain dense objects with overlapping image boundaries. We randomly select 20\% of data for testing, and the rest for training. Our network is implemented on the PyTorch framework and a Tesla P40 GPU with 24 GB memory is used.

Optimal Dataset Scale (ODS) and Optimal Image Scale (OIS) \cite{Alpher02} are two common used metrics in edge detection tasks. ODS uses a fixed threshold, which is calibrated globally for the whole dataset, to provide optimal performance. OIS evaluates the performance with the optimal threshold selected in a per-image basis. During quantitative evaluation, a maximum tolerance distance $d$ is used to match ground-truth edges, which is 0.0075 by default. In addition, to show the localization ability of different methods, we also leverage Dice Score Coefficient (DSC) and Hausdorff Distance (HD) as evaluation metrics. Notably, our proposed method can predict final binary edge maps without any post-processing. To be fair, for other compared methods, we evaluate the DSC and HD with a fixed optimal threshold calibrated globally for the whole dataset.

\begin{table*}[t]
	\setlength{\abovecaptionskip}{0pt}
	\setlength{\belowcaptionskip}{10pt}
	\renewcommand{\arraystretch}{1.0}
	\centering
	\caption{Evaluation of the effectiveness of hierachical boosting.} \label{table:StateofTheAat}
	\begin{tabular}{|c|c|c|c|c|c|c|c|c|c|c|}
		\hline
		{Metric}   & Side 1  & Side 2   & Side 3  & Side 4 & Side 5 & Side 6 & Side 7 & Side 8 & Fusion     \\ \hline
		
		{ODS (\%)}   &  65.78  &  68.08  &  73.53 &  80.44  &  81.50  &  82.83    &  84.49   & 84.94 & \textbf{85.05} \\ \hline
		
		{OIS (\%)}   &  66.10  &  68.20  & 73.60  &  80.44  &  81.50 &   82.90   &  84.60  & 84.97  & \textbf{85.05} \\ \hline
		
		{DSC (\%)}   & 50.21  & 50.19  & 50.19   & 50.20  & 50.22      & 50.22  & 51.75  & \textbf{51.24}  & 50.27    \\ \hline
		
		{HD}    & 12.832  & 12.832  & 12.832   & 12.832  &  12.826   & 12.815  & \textbf{12.088}  &  12.447   & 12.811  \\ \hline
	\end{tabular}
\end{table*}

\begin{table*}[t]
	\setlength{\abovecaptionskip}{0pt}
	\setlength{\belowcaptionskip}{10pt}
	\renewcommand{\arraystretch}{1.0}
	\centering
	\caption{Comparison with the state-of-the-art methods on 3D and 2D datasets.}\label{table:SOTA}
	\begin{tabular}{|c|c|c|c|c||c|c|c|c|}
		\hline
		\multirow{2}{*} & \multicolumn{4}{c||}{3D Brain Cell} & \multicolumn{4}{c|}{2D Nuclei}   \\ \cline{2-9} 
		& ODS (\%) &   OIS (\%)   & DSC (\%)  & HD  & ODS (\%) & OIS (\%) & DSC (\%) & HD  \\ \hline
		{HED \cite{xie15hed}} & 68.52    &  69.13    &   28.04   &  22.811 & 82.28  & 82.66   & 39.54   & 44.981    
		\\ \hline
		{RCF \cite{liu2019richer}} &  80.84   &  81.20   &  35.23   &  22.298  & 87.25  & 88.12  & 40.03   & 31.601 
		\\ \hline
		{BDCN \cite{he2019bdcn}} & \textbf{85.43} & \textbf{85.43} & 43.83 &  18.341 & 88.08  & \textbf{88.76}   & 50.76 &\textbf{23.340}
		\\ \hline
		{Proposed} & 85.05  & 85.05 &\textbf{50.27}  &\textbf{12.811} & \textbf{88.68}   & 88.68    & \textbf{51.65}  &  30.252 
		\\ \hline
	\end{tabular}
\end{table*}

\noindent\textbf{Ablation Study.} 
To evaluate the effectiveness of our proposed logical gate operation, we conduct a set of experiments with and without the logical gate. Table 1 shows that the logical gate improves the performance in the coarse phase distinctly, while the model cannot converge well without the logical gate in the refinement phase. With the logical gate in the refinement phase, the ODS, OIS, DSC and HD can be further improved to 85.05\%, 85.05\%, 50.27\% and 12.811, respectively. From the arrows in Fig. 5, we can see that the edge maps can be improved to be clearer and thinner with the logical gate on both 3D and 2D datasets.

We also evaluate the effectiveness of the deeply supervised hierarchical boosting in crisp edge refinement network. Each side is corresponding to an output shown in Fig. 3. The results in Table 2 show that the performance is in an increasing trend with the helpful hierarchical information. With all convolutional layers combined to employ ample features, it achieves a boost in performance.

\noindent\textbf{Comparisons with State of the Art.} We compare the proposed method with state-of-the-art methods including HED \cite{xie15hed}, RCF \cite{liu2017richer1,liu2019richer}, and BDCN \cite{he2019bdcn}, and conduct extensive experiments on those two datasets. The results shown in Table \ref{table:SOTA} indicate that the performance of our proposed method is comparable with BDCN, which is on the cutting edge. It is worth to note that BDCN, as well as all other compared methods, need an optimal threshold to get the binary edge map, while our proposed method can predict final edge map with crisp boundaries directly. Last but not least, the DSC of the proposed method outperforms all other methods, which indicates the superior ``correctness'' of our proposed method in distinguishing edge and non-edge pixels. Fig. 6 shows the edge detection results of different methods. As indicated by the arrows, the edge response of our proposed method is more precise, sharper and clearer than compared methods. 

\begin{figure}[t]
	\begin{center}
		\includegraphics[width=1\linewidth]{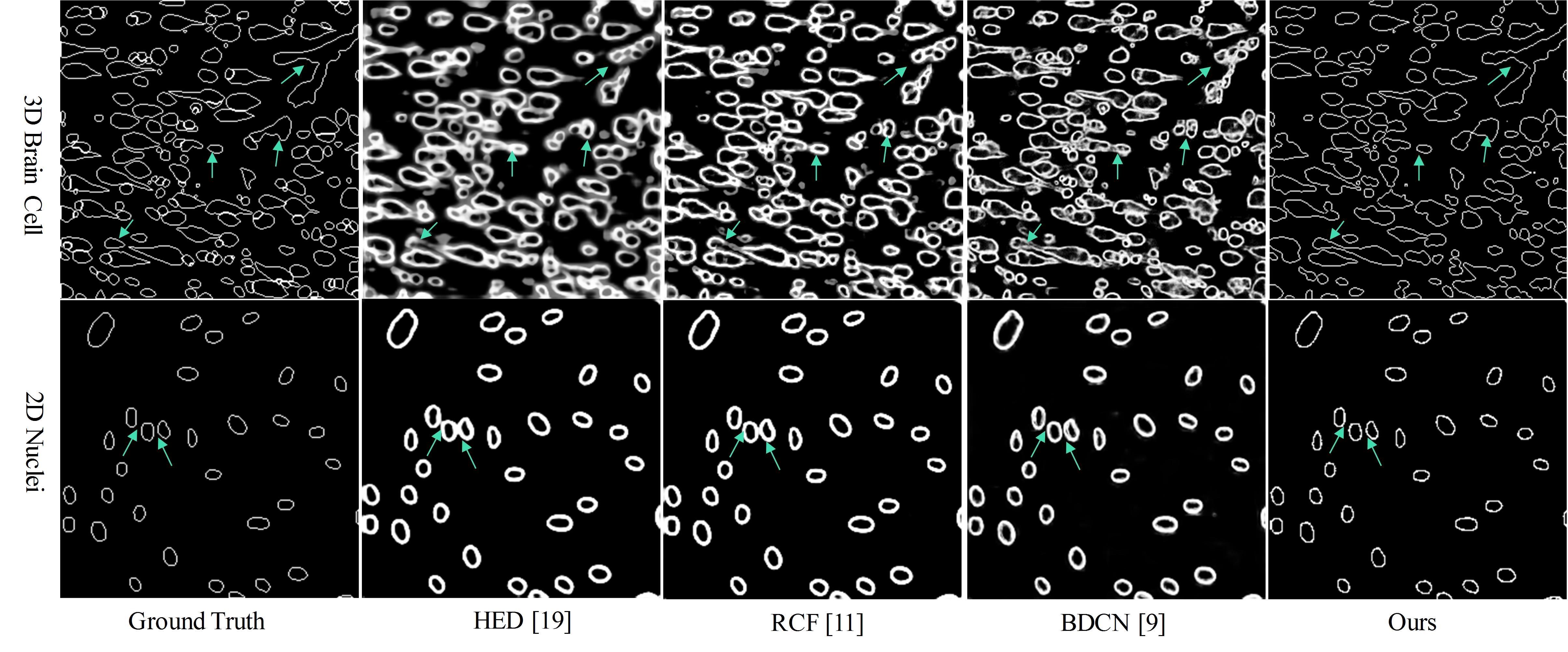}
	\end{center}
	\caption{The results of different edge detection methods.}
	\label{fig:long}
	\label{fig:onecol}
\end{figure}

\begin{figure}[t]
	\begin{center}
		\includegraphics[width=1\linewidth]{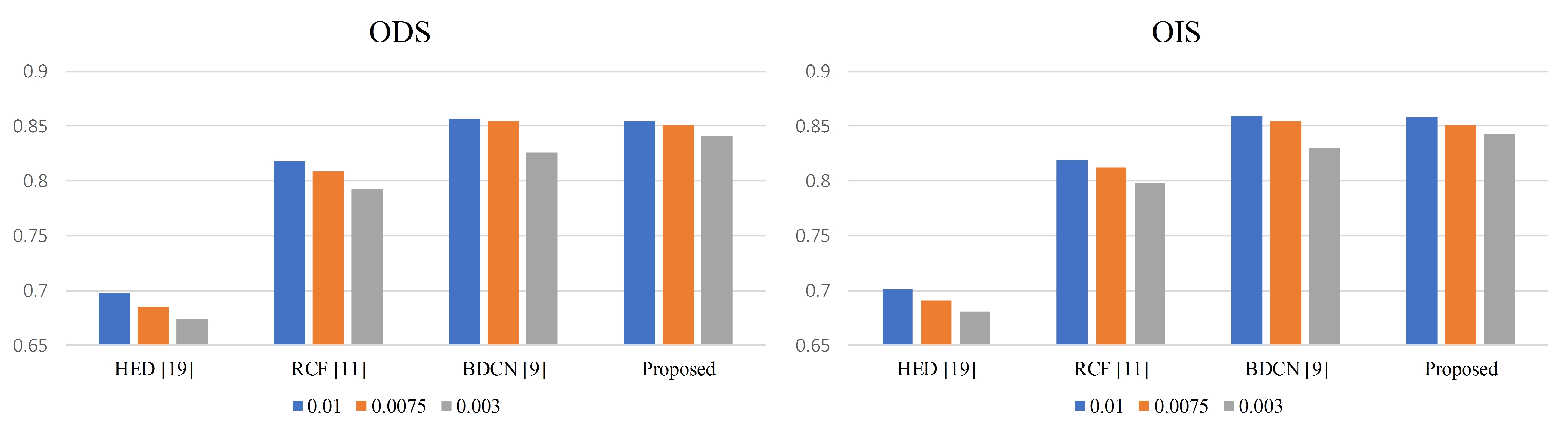}
	\end{center}
	\caption{Performance (ODS and OIS) as a function of the maximum tolerance distance for edge matching to evaluate the ``crispness'' of edges.}
	\label{fig:long}
	\label{fig:onecol}
\end{figure}

We further evaluate the ``crispness'' of edges from the proposed method by varying the maximum tolerance distance $d$ when matching ground-truth edges during evaluation, which is shown in Fig. 7. The performance of all methods decreases when tightening the evaluation criterion from 0.01 to 0.003. The performance of our proposed method decreases slowly when $d$ decreases. In contrast, the performance of compared methods drops quickly. In fact, the ODS gap between the proposed method and RCF increases from 3.68\% to 4.80\%, and the OIS gap increases from 3.87\% to 4.39\%. However, the gaps between BDCN and CRF become closer with a smaller $d$, decreasing from 3.85\% to 3.32\% in the ODS gap and from 3.94\% to 3.22\% in the OIS gap, respectively. The results suggest that our proposed method produces a crisp edge.

\section{Conclusion}
In this paper, we proposed a novel coarse-refine framework for crisp edge detection, called logical refinement network. Motivated by the logical relationship between binary segmentation and edge maps, we proposed a logical gate in which the segmentation and edge features are utilized to gradually improve the quality of edge maps. In conjunction with logical gate, the proposed method detects crisp and clear edges, which stick to the actual image boundaries. Experiments show that the proposed method is able to detect accurate and crisp 2D and 3D image edges, and it is also the first attempt to address the 3D edge detection problem with deep learning. In terms of the correctness and crispness of detected edges, our proposed method significantly outperforms existing state-of-the-art algorithms in both 2D and 3D. 

\section*{Acknowledegments}
This work was supported by the grants from Key Area Research and Development Program of Guangdong Province, China (No. 2018B010111001) and the Science and Technology Program of Shenzhen, China (No. ZDSYS201802021814180).
\bibliographystyle{splncs04}
\bibliography{egbib}

\begin{thebibliography}{10}
\providecommand{\url}[1]{\texttt{#1}}
\providecommand{\urlprefix}{URL }
\providecommand{\doi}[1]{https://doi.org/#1}

\bibitem{Alpher02}
Arbelaez, P., Maire, M., Fowlkes, C., Malik, J.: Contour detection and
  hierarchical image segmentation. IEEE Transactions on Pattern Analysis and
  Machine Intelligence  \textbf{33}(5),  898--916 (2010)

\bibitem{badrinarayanan2017segnet}
Badrinarayanan, V., Kendall, A., Cipolla, R.: Seg{N}et: A deep convolutional
  encoder-decoder architecture for image segmentation. IEEE Transactions on
  Pattern Analysis and Machine Intelligence  \textbf{39}(12),  2481--2495
  (2017)

\bibitem{Alpher14}
Bertasius, G., Shi, J., Torresani, L.: Deep{E}dge: A multi-scale bifurcated
  deep network for top-down contour detection. Proceedings of the IEEE
  Conference on Computer Vision and Pattern Recognition pp. 4380--4389 (2015)

\bibitem{Alpher17}
Bertasius, G., Shi, J., Torresani, L.: High-for-low and low-for-high: Efficient
  boundary detection from deep object features and its applications to
  high-level vision. Proceedings of the IEEE International Conference on
  Computer Vision pp. 504--512 (2015)

\bibitem{caicedo2019nucleus}
Caicedo, J.C., Goodman, A., Karhohs, K.W., Cimini, B.A., Ackerman, J.,
  Haghighi, M., Heng, C., Becker, T., Doan, M., McQuin, C., et~al.: Nucleus
  segmentation across imaging experiments: the 2018 data science bowl. Nature
  methods  \textbf{16}(12),  1247--1253 (2019)

\bibitem{Alpher10}
Canny, J.: A computational approach to edge detection. IEEE Transactions on
  Pattern Analysis and Machine Intelligence (6),  679--698 (1986)

\bibitem{dollar2013structured}
Doll{\'a}r, P., Zitnick, C.L.: Structured forests for fast edge detection.
  Proceedings of the IEEE International Conference on Computer Vision pp.
  1841--1848 (2013)

\bibitem{girshick2014rich}
Girshick, R., Donahue, J., Darrell, T., Malik, J.: Rich feature hierarchies for
  accurate object detection and semantic segmentation. Proceedings of the IEEE
  Conference on Computer Vision and Pattern Recognition pp. 580--587 (2014)

\bibitem{he2019bdcn}
He, J., Zhang, S., Yang, M., Shan, Y., Huang, T.: Bi-directional cascade
  network for perceptual edge detection. Proceedings of the IEEE Conference on
  Computer Vision and Pattern Recognition pp. 3828--3837 (2019)

\bibitem{hu2019dynamic}
Hu, Y., Chen, Y., Li, X., Feng, J.: Dynamic feature fusion for semantic edge
  detection. arXiv preprint arXiv:1902.09104  (2019)

\bibitem{liu2019richer}
Liu, Y., Cheng, M.M., Hu, X., Bian, J.W., Zhang, L., Bai, X., Tang, J.: Richer
  convolutional features for edge detection. IEEE Transactions on Pattern
  Analysis and Machine Intelligence  \textbf{41}(8),  1939--1946 (2019)

\bibitem{liu2017richer1}
Liu, Y., Cheng, M.M., Hu, X., Wang, K., Bai, X.: Richer convolutional features
  for edge detection. IEEE Conference on Computer Vision and Pattern
  Recognition pp. 3000--3009 (2017)

\bibitem{Alpher11}
Maire, M., Stella, X.Y., Perona, P.: Reconstructive sparse code transfer for
  contour detection and semantic labeling. Asian Conference on Computer Vision
  pp. 273--287 (2014)

\bibitem{Alpher04}
Pont-Tuset, J., Arbelaez, P., Barron, J.T., Marques, F., Malik, J.: Multiscale
  combinatorial grouping for image segmentation and object proposal generation.
  IEEE Transactions on Pattern Analysis and Machine Intelligence
  \textbf{39}(1),  128--140 (2016)

\bibitem{Qin_2019_CVPR}
Qin, X., Zhang, Z., Huang, C., Gao, C., Dehghan, M., Jagersand, M.: {BASN}et:
  Boundary-aware salient object detection. The IEEE Conference on Computer
  Vision and Pattern Recognition  (June 2019)

\bibitem{ronneberger2015u}
Ronneberger, O., Fischer, P., Brox, T.: U-{Ne}t: Convolutional networks for
  biomedical image segmentation. International Conference on Medical Image
  Computing and Computer Assisted Intervention pp. 234--241 (2015)

\bibitem{Alpher15}
Shen, W., Wang, X., Wang, Y., Bai, X., Zhang, Z.: Deep{C}ontour: A deep
  convolutional feature learned by positive-sharing loss for contour detection.
  Proceedings of the IEEE Conference on Computer Vision and Pattern Recognition
  pp. 3982--3991 (2015)

\bibitem{Alpher16}
Wang, Y., Zhao, X., Huang, K.: Deep crisp boundaries. Proceedings of the IEEE
  Conference on Computer Vision and Pattern Recognition pp. 3892--3900 (2017)

\bibitem{xie15hed}
Xie, S., Tu, Z.: Holistically-nested edge detection. Proceedings of the IEEE
  International Conference on Computer Vision pp. 1395--1403 (2015)

\bibitem{Alpher20}
Yu, Z., Feng, C., Liu, M.Y., Ramalingam, S.: Case{N}et: Deep category-aware
  semantic edge detection. Proceedings of the IEEE Conference on Computer
  Vision and Pattern Recognition pp. 5964--5973 (2017)

\bibitem{yu2018seal}
Yu, Z., Liu, W., Zou, Y., Feng, C., Ramalingam, S., Vijaya~Kumar, B., Kautz,
  J.: Simultaneous edge alignment and learning. European Conference on Computer
  Vision  (2018)

\bibitem{zimmermann2019faster}
Zimmermann, R.S., Siems, J.N.: Faster training of {M}ask {R-CNN} by focusing on
  instance boundaries. Computer Vision and Image Understanding  \textbf{188},
  102795 (2019)

\end{thebibliography}

\end{document}